\documentclass[10pt,letterpaper]{article}

\usepackage{cogsci}

\usepackage{times}
\usepackage{latexsym}
\usepackage{graphicx}
\usepackage{booktabs}
\usepackage{multirow}
\usepackage{amsmath}
\usepackage{multirow}
\usepackage{makecell}

\cogscifinalcopy 

\usepackage[
  style=apa,
  natbib=true,
  annotation=false,
]{biblatex}
\usepackage{float} 

\title{Debiasing Central Fixation Confounds Reveals a Peripheral “Sweet Spot” for Human-like Scanpaths in Hard-Attention Vision}

\author[1]{\mbox{Pengcheng Pan}}
\author[1]{\mbox{Yonekura Shogo}}
\author[1]{\mbox{Yasuo Kuniyoshi}}
\affil[1]{Department of Mechano-Informatics, The University of Tokyo}

\begin{document}

\maketitle

\begin{abstract}
	Human eye movements in visual recognition reflect a balance between foveal sampling and peripheral context, but evaluating whether artificial scanpaths are "human-like" is difficult on object-centric datasets with strong center bias.
	Using Gaze-CIFAR-10, we show that a trivial \emph{center-fixation baseline} achieves surprisingly strong scores under common scanpath metrics, blurring the distinction between behavioral alignment and central tendency.
	We introduce \textbf{GCS} (Gaze Consistency Score), a practical center-debiased and movement-aware score that normalizes against human and corner references, subtracts the center baseline, and adds a small movement-similarity term.
	Applying GCS to a hard-attention classifier under varied fovea--periphery constraints identifies a restricted mid-range regime: a moderate foveal patch with peripheral context yields stronger center-debiased alignment than either narrower or broader alternatives.
	This regime is not identified by accuracy alone; the highest-accuracy setting differs from the best-GCS setting.
	These results highlight the need for bias-aware scanpath evaluation and suggest that, on Gaze-CIFAR-10 and under this hard-attention setting, perceptual constraints shape when task-trained policies appear relatively human-like.
\end{abstract}


\section{Introduction}

\begin{figure}[t]
	\centering
	\includegraphics[width=\columnwidth]{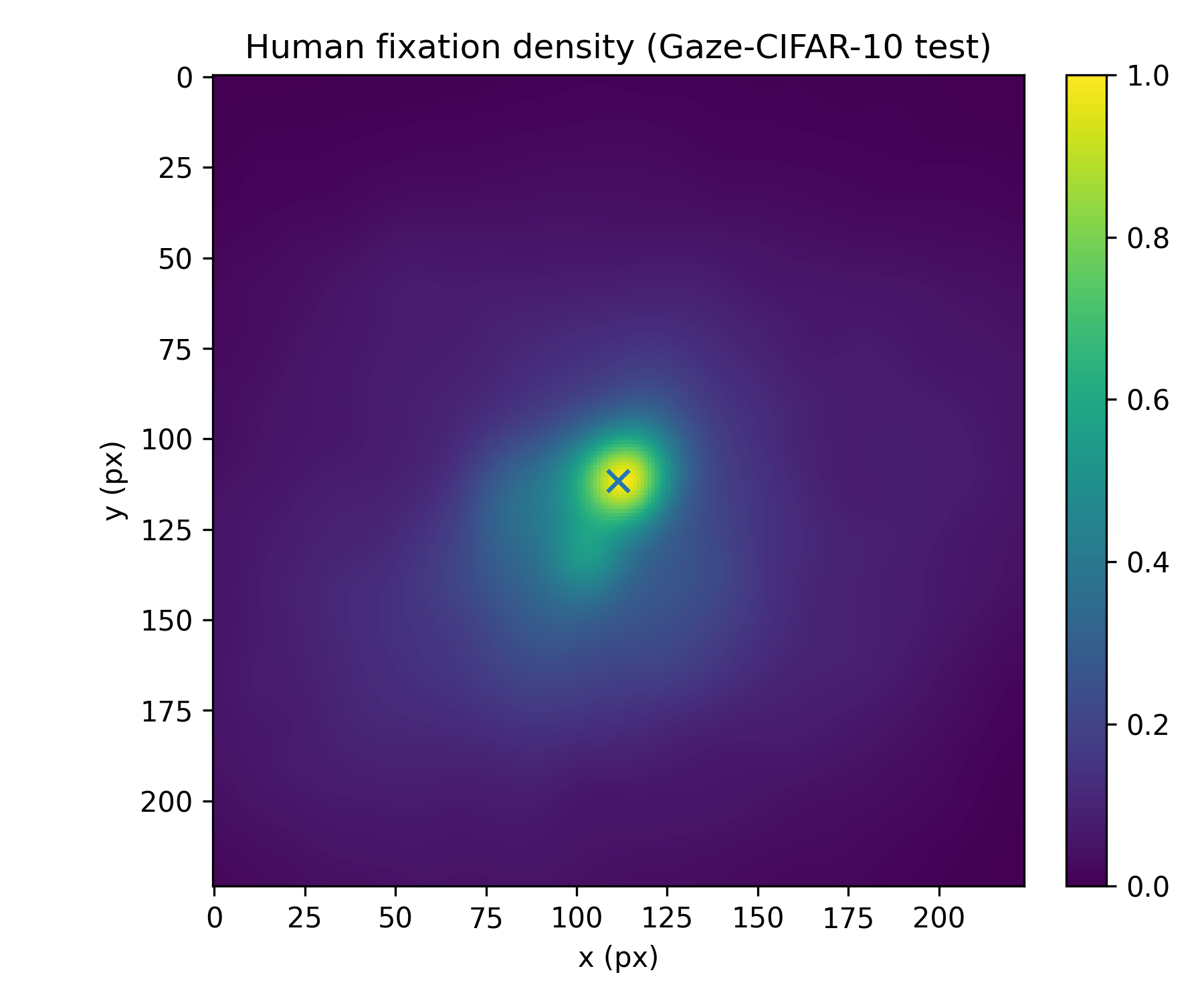}\\
	\vspace{2mm}
	\includegraphics[width=\columnwidth]{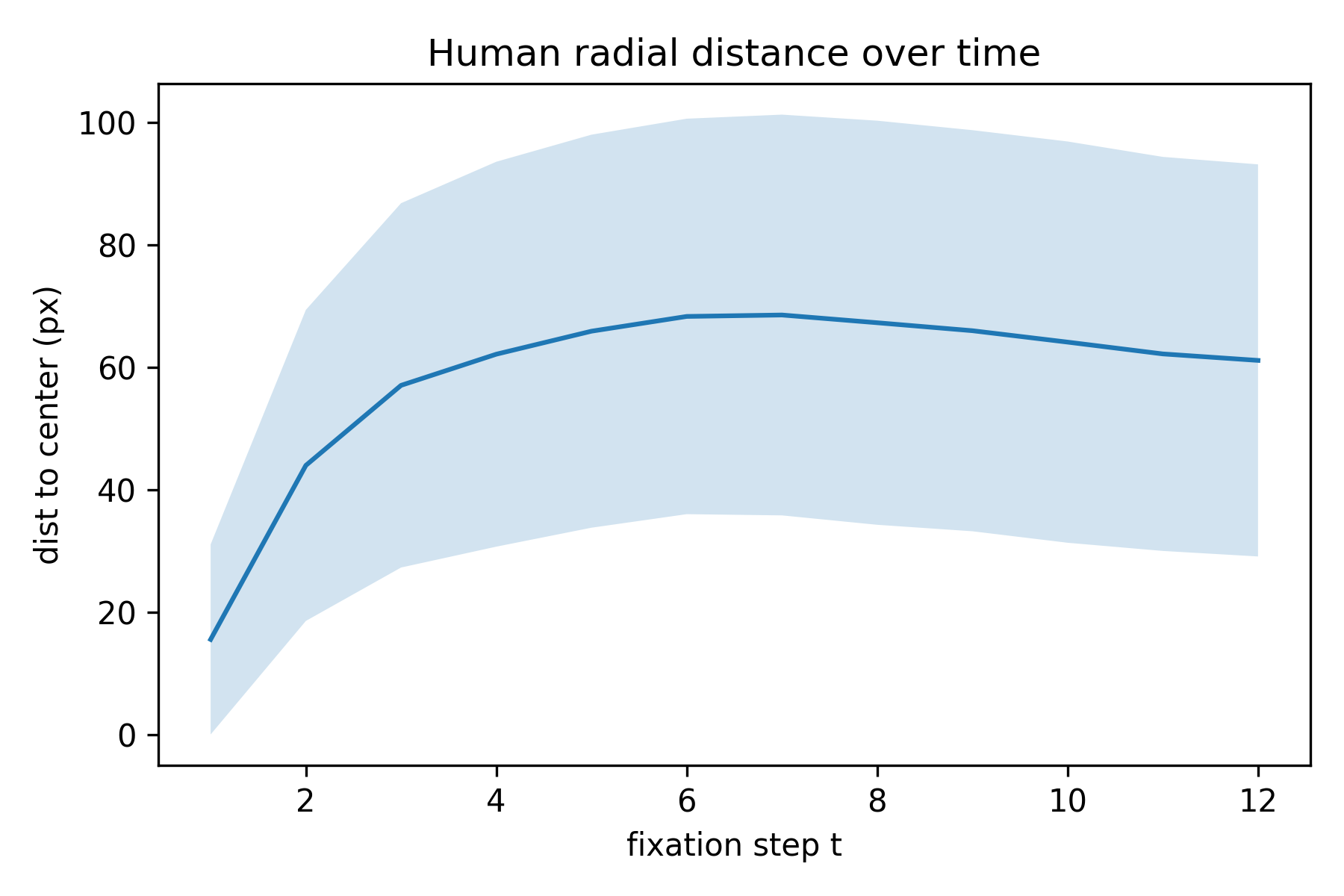}
	\caption{Human fixation density and radial distance over time (Gaze-CIFAR-10 test). The distribution is strongly center-biased, which can inflate scanpath similarity for trivial center policies.}
	\label{fig:centerbias}
\end{figure}
Human visual perception is an active process.
Rather than processing the entire visual field uniformly, humans rely on a foveated visual system and sequential eye movements to gather task-relevant information.
Classic work by \cite{yarbus1967} demonstrated that eye movement patterns are systematically shaped by task demands, suggesting that scanpaths can provide a window into underlying cognitive strategies.

In recent years, this idea has motivated the development of \emph{hard-attention} models in machine vision, which explicitly select where to look at each step while performing tasks such as image classification or visual search \citep{mnih2014ram, ba2015dram,pan2025mram}.
These models offer a promising computational framework for studying perception under resource constraints.
However, an open question remains: \emph{when do the scanpaths produced by such models trained purely on the task meaningfully resemble human eye movements?}

A major challenge in answering this question lies in how scanpaths are evaluated.
A wide range of metrics have been proposed to compare fixation sequences, made available on FixaTons tools \citep{FixaTons}, including Dynamic Time Warping (\citealp{DTW}), ScanMatch \citep{SM}, Normalized Scanpath Saliency (\citealp{NSS}), and AUC-based measures originally developed for saliency evaluation \citep{judd2012}.
These metrics capture complementary aspects of scanpaths, such as temporal alignment, spatial overlap, and distributional similarity.

However, a long-standing finding in human eye-movement research complicates the interpretation: \emph{human fixations are strongly biased toward the image center}.
This \textbf{central fixation bias} has been robustly documented across free viewing and task-driven settings \citep{tatler2007centerbias, tatler2011bias, bindemann2010}.
Proposed explanations include photographer bias, experimental framing, oculomotor constraints, and learned expectations about where informative content is likely to appear \citep{smith2013centerbias}.
Crucially, center bias is not merely noise, it is a systematic property of many gaze datasets.

\paragraph{Center bias as a confound in scanpath evaluation.}
When center bias is strong, scanpath similarity metrics can be dominated by marginal spatial distributions rather than by genuine strategy similarity.
Indeed, prior work in saliency modeling has shown that center-biased predictors can achieve high AUC and NSS scores without modeling image content \citep{borji2013, borji2019}.
Despite this, task-driven scanpath evaluation has rarely applied explicit chance correction or strong spatial baselines.

This issue is particularly acute for object-centric datasets, where objects are frequently centered by design.
In such settings, a trivial strategy that repeatedly fixates the center may appear "human-like" under standard scanpath metrics, even if it lacks meaningful temporal structure or task-driven exploration.
As a result, high scanpath similarity does not necessarily imply cognitively plausible attention control.

\paragraph{Active vision, perceptual constraints, and strategy regimes.}
Classic work in psychology and neuroscience emphasizes that scanpaths depend strongly on task demands and internal goals \citep{yarbus1967}.
In cognitive robotics and AI, the active perception framework formalizes perception as an action-conditioned process \citep{Bajcsy1988,Ballard1991},
motivating models that must decide \emph{where to look} under limited sensing.

Normative models of active vision show that optimal fixation selection depends critically on sensory constraints such as acuity, noise, and field-of-view \citep{najemnik2005, geisler2008}.
Manipulating these constraints can induce qualitative shifts in behavior, from exploratory search to more conservative sampling.

Similar trade-offs arise in artificial agents.
Hard-attention models with access to broader spatial context can reduce the need for movement, sometimes relying on shortcut strategies that bypass sequential evidence accumulation \citep{geirhos2020shortcut}.
Conversely, overly narrow views can force excessive exploration.
These observations motivate a scoped hypothesis: under this hard-attention setting, relatively human-like scanpaths may appear most clearly within a restricted regime of perceptual constraints.

\paragraph{Goals and contributions.}
In this work, we study these issues using Gaze-CIFAR-10 \citep{gazecifar}, a gaze-annotated object recognition dataset.
We make three contributions:

\begin{enumerate}
	\item \textbf{Quantifying strong center bias in Gaze-CIFAR-10:} We provide direct empirical evidence that Gaze-CIFAR-10 exhibits strong center bias, and show that a simple center-fixation baseline achieves surprisingly high scanpath similarity under DTW, ScanMatch, NSS, and AUC.
	\item \textbf{A center-debiased, movement-aware evaluation score.}
	We introduce \textbf{GCS} (Gaze Consistency Score), which (i) normalizes similarity using human--human agreement as an upper bound and a corner baseline as a lower bound, and (ii) explicitly subtracts a center baseline to reduce dataset-induced bias, thereby emphasizing movement dynamics beyond mere spatial overlap.
	\item \textbf{Peripheral sweet spot as a mechanistic regime:} Using GCS, we identify a restricted, setting-dependent regime in a hard-attention classifier: under Gaze-CIFAR-10, a moderate foveal patch with peripheral context yields the strongest center-debiased alignment. We interpret this as a mechanistic regime analysis rather than a strict human viewport matching.
\end{enumerate}

Together, these results highlight the importance of bias-aware evaluation when using scanpaths to evaluate relative human-likeness, and suggest that perceptual constraints can shape artificial eye movement strategies in this object-centric benchmark.

\section{Method}
\subsection{Gaze-CIFAR-10}
We use the Gaze-CIFAR-10 dataset, which provides human fixation sequences collected during CIFAR-10 image recognition.
We evaluate on the test split ($N=9{,}116$ images). We use 12 fixation steps for both human and model scanpaths so that all metrics compare trajectories under the same temporal budget.

\subsection{Hard-Attention Classifier Under Constrained Vision}
Our model is a hard-attention recurrent vision agent trained for image classification.
To implement the above \emph{constrained-vision} classifier as an \emph{active perceiver}, we use a Multi-Level Recurrent Attention Model (MRAM), a hierarchical recurrent agent that couples \emph{eye-movement control} with \emph{evidence accumulation}.
Conceptually, MRAM can be read as an idealized perception--action loop under partial observability: at each time step $t$, the agent (i) fixates, (ii) receives a limited retinal sample (a "glimpse") determined by a \emph{foveal patch size}, (iii) updates an internal state, and (iv) chooses where to look next.

\paragraph{Retina-like sensing as momentary observation.}
Given the current fixation location $l_{t-1}$, a glimpse module extracts a high-resolution foveal crop and, when enabled in specific setting, a wider low-resolution peripheral crop.
The peripheral crop is down-sampled to the same spatial size as the foveal crop, and the foveal and peripheral channels are concatenated before being embedded into a sensory vector $g_t$.
Thus the peripheral channel carries \emph{wider field-of-view but lower acuity} information, whereas the foveal channel carries \emph{higher acuity but narrower} information.
This provides a simple computational analogue of the fovea--periphery trade-off familiar from human vision.

\paragraph{Two recurrent levels and training}
MRAM maintains two recurrent states updated sequentially at each time step:
(1) a \emph{lower-level} recurrent state $h_t^{(1)}$ that is driven directly by the current sensory sample $g_t$, and
(2) a \emph{higher-level} recurrent state $h_t^{(2)}$ that integrates information over time and supports the final decision.
Intuitively, $h_t^{(1)}$ plays the role of a fast, action-oriented sensorimotor state (what to do next), whereas $h_t^{(2)}$ acts like a slower, more abstract state that accumulates evidence (what is it).
This separation mirrors classic cognitive distinctions between \emph{sampling and eye movement control} and \emph{recognition in cortex}, and is consistent with hierarchical views of active perception in which lower circuitry drives saccade-like control while higher circuitry stabilizes task-relevant representations.

The higher-level state $h_t^{(2)}$ is mapped to a categorical prediction over labels, yielding a running belief that can be updated as new glimpses arrive.
We take the final-step prediction as the agent's classification decision after a fixed sampling budget at 12 steps, which corresponds to a bounded evidence-accumulation process.
From the lower-level state $h_t^{(1)}$, a stochastic location policy samples the next fixation $l_t$.
In our implementation, $l_t$ is a 2D coordinate (normalized to the image plane), so the observable output is a scanpath-like sequence of fixations.

Because fixation selection is a discrete action, MRAM is trained with a REINFORCE algorithm (a Reinforcement Learning method) so that successful classifications reinforce the sampling strategies that produced informative glimpses.

\paragraph{Parameters to test our hypothesis.}
We hypothesize that relatively human-like scanpaths are most likely when the model must solve the same classification task under constrained information-processing conditions. 
In particular, we expect human-like scanpaths to emerge most strongly under a moderately constrained field of view: if the visual input is too restricted, the agent lacks sufficient context to guide efficient sampling; if it is too broad, the task can be solved via passive, shortcut-like recognition with little need for active exploration.

We study three sensory configurations that vary peripheral context:

\begin{itemize}
	\item \textbf{fov\_only:} observes only the foveal crop.
	\item \textbf{fov+per:} concatenates the foveal crop with a moderate low-resolution peripheral context.
	\item \textbf{big\_per:} uses a wider peripheral context, increasing the effective field-of-view and reducing the need for active movement.
\end{itemize}

We sweep foveal patch size in $\{2,4,6,8,10,12,14,16,20,24,28,32\}$ pixels, keeping the number of steps and the main sweep seed fixed.
Intuitively, patch size corresponds to an \emph{effective viewing condition hypothesis}: smaller patches require more active exploration, while larger patches allow more passive recognition.

\subsection{Standard Scanpath Metrics}
We report four commonly used scanpath similarity metrics: 
\textbf{DTW} (lower is better), \textbf{ScanMatch} (higher is better), 
\textbf{NSS} (higher is better), and \textbf{AUC} (higher is better).

To contextualize metric values under dataset bias, we compute three references:
(i) an \textbf{upper bound} by comparing each human scanpath to itself (identical sequence),
(ii) a \textbf{lower bound} using a \textbf{corner-fixation} policy,
and (iii) a \textbf{center baseline} using an always-center policy.
The center baseline is a fixed policy that repeats the image-center fixation for all 12 steps. The corner baseline repeats a corner fixation for all 12 steps; we use it as a conservative spatially mismatched lower reference, not as a theoretical worst possible policy.

\subsection{GCS: Center-Debiased, Movement-Aware Score}
\begin{figure}[t]
	\centering
	\includegraphics[width=\columnwidth]{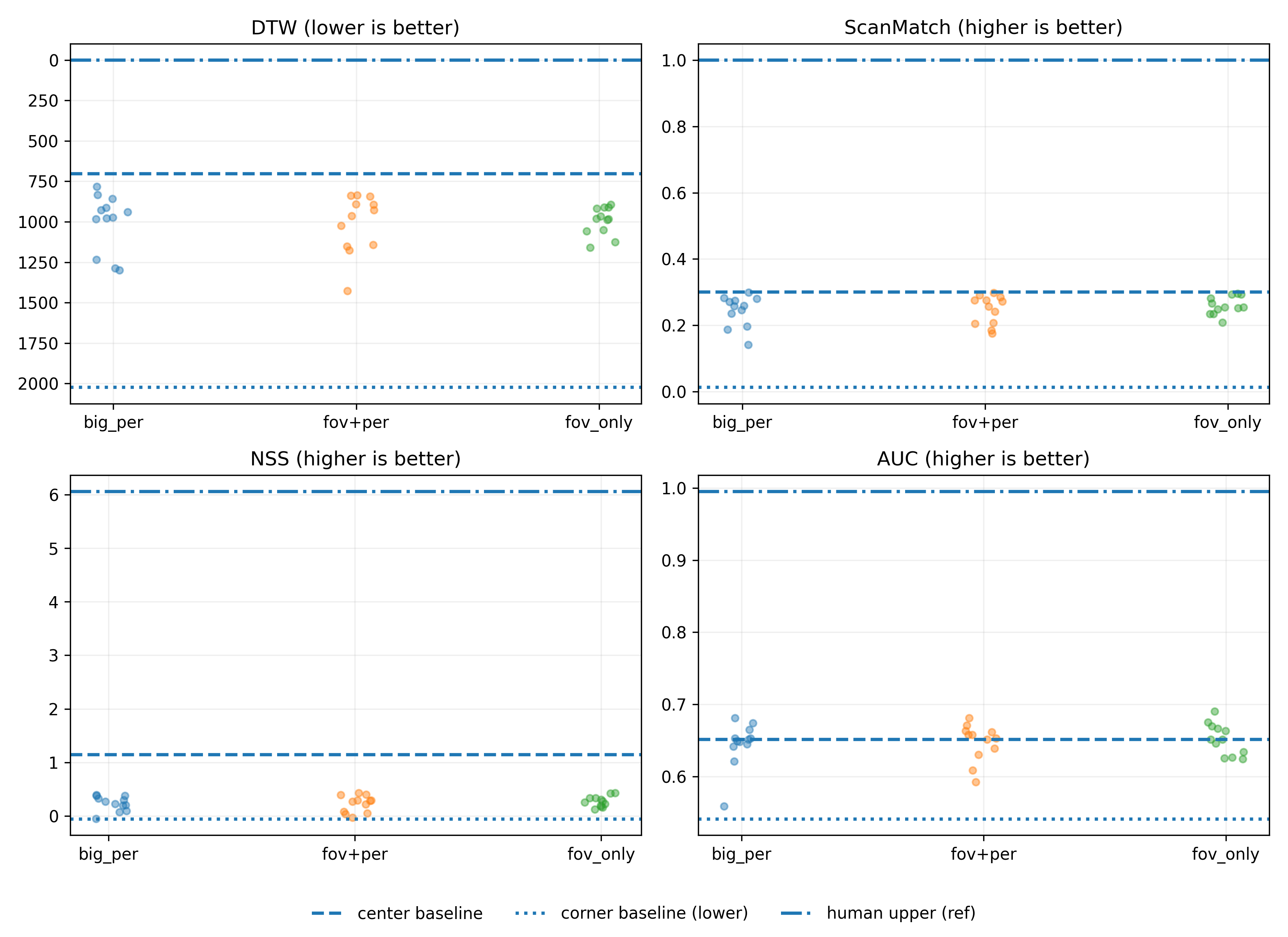}
	\caption{Raw scanpath metric distributions for three sensory settings. Dashed lines: center baseline; dotted lines: corner lower bound; dash-dot lines: human upper bound. Standard metrics can be optimistic due to center bias.}
	\label{fig:metricbaseline}
\end{figure}

We define \textbf{GCS} (\emph{Gaze Consistency Score}) to reduce center-bias inflation.
GCS is designed to answer a conservative question: does a learned policy exceed what can be explained by the dataset's center prior? First, each raw metric is normalized between a human reference and a spatially mismatched corner reference. Second, the normalized center baseline is subtracted, so policies receive credit only for performance above an always-center strategy. Third, we add a small movement-similarity term to discourage policies that match human fixation density only through central overlap while lacking human-like trajectory dynamics.
For each metric $M$, we normalize scores to $[0,1]$ using the upper/lower bounds and then subtract the center baseline:
\begin{align}
	\tilde{D} &= \frac{D_{\min}-D}{D_{\min}-D_{\max}} \quad (\text{DTW})\\
	\tilde{M} &= \frac{M - M_{\min}}{M_{\max}-M_{\min}} \quad (\text{SM,NSS,AUC})\\
	\tilde{M}_{db} &= \tilde{M} - \tilde{M}_{center}.
\end{align}
We compute a relative-error distance:
$d=\sqrt{\frac{1}{K}\sum_{k=1}^{K}\left(\frac{|f_k^{\text{model}}-f_k^{\text{human}}|}{|f_k^{\text{human}}|+\epsilon}\right)^2}$ where $f_k$
denotes a run-level movement statistic including total path, mean saccade amplitude, mean distance-to-center, spatial coverage, direction entropy, and collapse rate,
and map it to $\mathrm{Sim}_{move}=\exp(-d/\tau)$.
Finally, we add the movement term $\mathrm{Sim}_{move}$ to discourage policies that match only spatial center overlap but not temporal dynamics:
\begin{equation}
	\mathrm{GCS} = \frac{1}{4}\sum_{M \in \{D,SM,NSS,AUC\}} \tilde{M}_{db} \;+\; \lambda \, \mathrm{Sim}_{move},
\end{equation}
with $\lambda=0.1$.
GCS can be negative. A negative value indicates that, after normalization and center-baseline subtraction, the policy falls below the center-calibrated reference overall.
Sensitivity analysis shows that varying $\lambda$ (0 to 0.5) does not change the qualitative setting-dependent pattern, indicating that our conclusions are driven by debiased alignment rather than a particular weighting of movement similarity.
To make this confound explicit for model evaluation, we compare learned policies against the three calibration references under each metric (Figure~\ref{fig:metricbaseline}).
Across DTW/ScanMatch/AUC, many learned policies lie close to the center baseline, supporting the need for debiasing.

\begin{table}[t]
	\centering
	\begin{tabular}{lcccc}
		\toprule
		\textbf{Baseline} & \textbf{DTW} $\downarrow$ & \textbf{SM} $\uparrow$ & \textbf{NSS} $\uparrow$ & \textbf{AUC} $\uparrow$\\
		\midrule
		\makecell[lt]{Identical-Human\\(upper bound)} & $0.003$ & $1.000$ & $6.052$ & $0.995$ \\
		\makecell[lt]{Corner-Human\\(lower bound)} & $2023.87$ & $0.013$ & $-0.053$ & $0.541$ \\
		\makecell[lt]{Center-Human\\(baseline)} & $702.24$ & $0.300$ & $1.145$ & $0.6515$ \\
		\bottomrule
	\end{tabular}
	\caption{Calibration baselines on Gaze-CIFAR-10 test. The center baseline is unexpectedly strong under standard scanpath metrics.}
	\label{tab:baselines}
\end{table}

\section{Results}
\subsection{Center Bias Inflates Standard Scanpath Similarity on Gaze-CIFAR-10}
We first report a confound that is critical for interpreting behavioral alignment on object-centric datasets.
Figure~\ref{fig:centerbias} shows that human fixations are strongly center-weighted over time.
Consistent with this bias, Table~\ref{tab:baselines} shows that a trivial center-fixation policy already attains high similarity under common metrics.
As a consequence, raw scanpath metrics can substantially overestimate alignment for policies that exploit the dataset's center prior.

\subsection{Peripheral Sweet Spot Revealed by GCS}
\begin{figure}[t]
	\centering
	\includegraphics[width=\columnwidth]{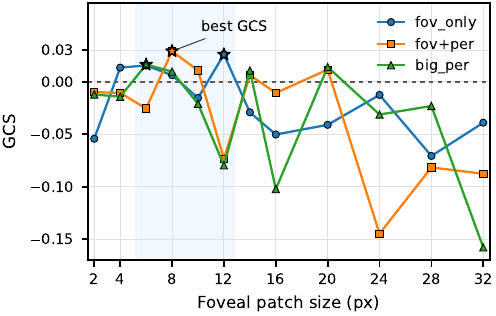}
	\caption{GCS as a function of foveal patch size under three sensory settings. The horizontal line marks the center-calibrated reference (GCS $=0$). The sweet spot is setting-dependent: moderate fovea-plus-periphery reaches the strongest overall debiased alignment, while fov\_only shifts its optimum toward a larger patch size. This suggests that peripheral context changes the movement--evidence trade-off under constrained vision.}
	\label{fig:gcs}
\end{figure}

Figure~\ref{fig:gcs} plots GCS across patch sizes and sensory settings.
The strongest center-debiased alignment occurs in a restricted mid-range regime: \textbf{fov+per with patch size 8} achieves the highest GCS.
This point is not the highest-accuracy model. The highest accuracy is obtained by \textbf{fov\_only with patch size 8}, whose GCS is substantially smaller.
Notably, the peak is not identical across sensory settings: fov+per reaches its strongest debiased alignment at a smaller/intermediate patch size, fov\_only shifts its optimum toward a larger patch size, and big\_per shows a different profile under broader peripheral context.
This shift is consistent with the interpretation that peripheral context changes the effective information available per glimpse, but it is an interpretation of the observed regime shift rather than a direct measurement of information content.
Thus, the best recognition policy is not necessarily the most human-like under center-debiased scanpath evaluation.

The best overall alignment is achieved by \textbf{fov+per with patch size 8}:
\[
\begin{aligned}
	\mathrm{GCS}&=0.0291,\quad \mathrm{Acc}=58.50\%,\quad \mathrm{DTW}=835.5,\\
	\mathrm{SM}&=0.298,\quad \mathrm{NSS}=0.401\quad \mathrm{AUC}=0.681.
\end{aligned}
\]
In contrast, the highest accuracy configuration is \textbf{fov\_only with patch size 8}
(Acc $=65.31\%$), but its GCS is much smaller (GCS $=0.0060$), illustrating a non-trivial
trade-off between task performance and behavioral alignment.

\begin{figure}[t]
	\centering
	\includegraphics[width=\columnwidth]{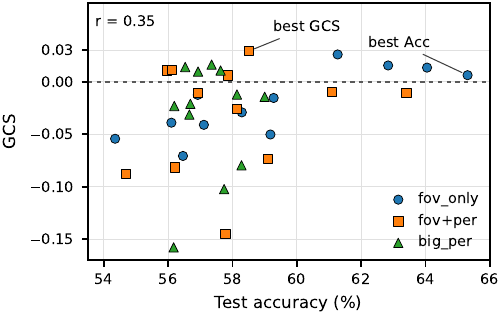}
	\caption{Accuracy versus GCS across sensory settings and patch sizes. Recognition performance and center-debiased scanpath alignment are only modestly coupled: the highest-accuracy configuration is not the best-GCS configuration.}
	\label{fig:accgcs}
\end{figure}

Table~\ref{tab:best} summarizes key regimes, including best-GCS and best-accuracy configurations for each sensory setting, as well as illustrative regimes that appear strong under raw metrics.
Figure~\ref{fig:accgcs} shows a modest correlation between recognition accuracy and scanpath similarity. Because the hard-attention agent is trained only for classification, this suggests that task learning can encourage the model to acquire some human-relevant evidence, but recognition success alone does not determine whether the scanpath is human-like after center-bias correction.

\begin{table*}[t]
	\centering
	\scriptsize
	\setlength{\tabcolsep}{4pt}
	\begin{tabular}{lccc|cccc|cc}
		\toprule
		Setting & Viewport & ps & Acc(\%) & DTW$\downarrow$ & SM$\uparrow$ & NSS$\uparrow$ & AUC$\uparrow$ & GCS$\uparrow$ & Path(px)\\
		\midrule
		Best raw DTW/SM & big\_per & 20 & 56.53 & 783.6 & 0.299 & 0.396 & 0.653 & 0.011 & 123.7\\
		Best raw AUC & fov\_only & 12 & 61.27 & 917.2 & 0.293 & 0.435 & 0.690 & 0.026 & 210.2\\
		\midrule
		Best Acc            & fov\_only & 8  & 65.31  & 986.9 & 0.281 & 0.427 & 0.669 & 0.006 & 329.8\\
		Best Acc (big\_per) & big\_per & 4    & 59.00  & 964.2 & 0.257 & 0.301 & 0.653 & -0.015 & 284.3\\
		Best Acc (fov+per)  & fov+per & 4     & 63.41  & 964.2 & 0.272 & 0.291 & 0.651 & -0.010 & 586.4 \\
		\midrule
		Best GCS & fov+per & 8 & 58.50 & 835.5 & 0.298 & 0.401 & 0.681 & 0.029 & 190.5\\
		Best GCS (big\_per) & big\_per & 6 & 57.35 & 857.5 & 0.283 & 0.378 & 0.673 & 0.016 & 158.4\\
		Best GCS (fov\_only; = Best AUC) & fov\_only & 12 & 61.27 & 917.2 & 0.293 & 0.435 & 0.690 & 0.026 & 210.2\\
		\bottomrule
	\end{tabular}
	\caption{Key regimes illustrating the center-bias pitfall of standard scanpath metrics and the debiased composite metric (GCS). ps: foveal patch size. Path: mean total path length over 12 glimpses.}
	\label{tab:best}
\end{table*}

\subsection{Movement Similarity at the Sweet Spot}
A possible alternative explanation is that the sweet spot simply reflects stronger central tendency.
The movement statistics argue against this interpretation.
The best-GCS run (fov+per, ps=8) still differs substantially from human scanpaths in absolute terms: total path is $190.5$ px vs.\ human $429.9$ px, and coverage is $7.37$ vs.\ $10.15$, indicating under-exploration.
However, compared with other regimes, it better balances center distance, spatial coverage, direction entropy, and collapse behavior.
Thus, the sweet spot should be interpreted as relative human-likeness under center-debiased evaluation, not as a full match to human scanpaths.

\subsection{Evidence Accumulation vs Movement}
To connect scanpaths to decision dynamics, we compare total movement with evidence accumulation, measured as the AUC of the true-class probability over glimpse steps.
Very small foveal windows often induce large movements, but this movement does not necessarily translate into efficient evidence accumulation.
Conversely, very large fields-of-view can reduce the need for movement, suggesting a shortcut-like regime under this setting.
The best-GCS regime lies between these extremes, consistent with a balance between active sampling and evidence gain.

\begin{figure}[t]
	\centering
	\includegraphics[width=\columnwidth]{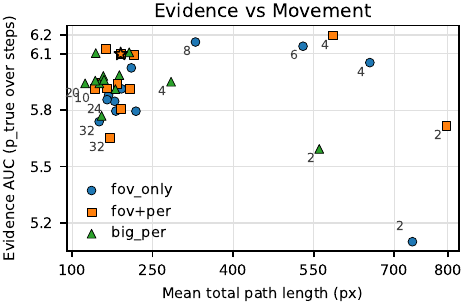}
	\caption{Evidence accumulation versus movement. Each point denotes one sensory configuration and foveal patch size, with numbers indicating patch size. Evidence is measured as the AUC of the true-class probability over glimpse steps. Very small windows induce large movements without proportional evidence gain, whereas overly broad views reduce the need for active sampling. The sweet-spot regime balances movement and evidence accumulation.}
	\label{fig:evidence}
\end{figure}

\section{Discussion}
\subsection{Does "Human-Like" Reduce to Matching Human Viewport?}
A reasonable concern is that changing patch size alters the agent's effective viewing geometry, whereas the human scanpaths were recorded under a fixed display setup.
Accordingly, our analysis varies the model's sensory constraint while the human recording geometry is fixed. The result should not be read as estimating the human viewport. Instead, it tests how active-vision policies change when sensory constraints are manipulated.

Crucially, our claim is not that a particular patch size \emph{matches} humans, but that three coupled observations hold:
(i) widely used scanpath metrics are sufficiently center-biased that trivial center-fixation policies can obtain high scores;
(ii) once we explicitly calibrate against such trivial baselines, only a narrow range of sensory configurations remains meaningfully above the center baseline; and
(iii) within this regime, movement statistics become more human-like displacement patterns in ways that cannot be explained by "being centered" alone.

This perspective yields a concrete, testable prediction for future data collection.
If the human viewing geometry is manipulated (e.g., viewing distance or effective image scale), human scanpaths should shift systematically across regimes that mirror our patch-size manipulation.
In particular, one may observe \emph{multiple} "sweet spots" across viewing conditions, rather than a single universal setting that defines human-likeness.

\subsection{Implications for Gaze Benchmarks and Cognitive Modeling}
For cognitive modeling, the results highlight a computational interpretation of gaze as \emph{policy under sensory constraints}.
For benchmark design, they caution that object-centric datasets can make scanpath evaluation deceptively easy.
We advocate reporting center and corner baselines by default, and using debiased metrics such as GCS when scanpaths are used as a claim of human-likeness.

\subsection{Limitations}
Our experiments focus on an object-centric, low-resolution classification benchmark, a single hard-attention architecture family, and a fixed 12-step horizon.
The main sweep uses a fixed seed; selected additional-seed runs provide only a limited sanity check, not a systematic multi-seed robustness analysis.
Therefore, the sweet-spot result should be interpreted as a mechanistic hypothesis in this setting, not as a general law of human gaze control.
Future work should test richer datasets, visual search tasks, additional architectures, and systematic human viewing manipulations.

\section{Conclusion}
We showed that Gaze-CIFAR-10 exhibits strong center bias, such that a trivial center-fixation baseline achieves high scores under common scanpath metrics.
We introduced GCS, a practical center-debiased and movement-aware evaluation score that calibrates model scanpaths against human, corner, and center references.
Under this metric, a hard-attention classifier shows a restricted mid-range regime in which foveal and peripheral information jointly yield stronger center-debiased alignment than either overly narrow or overly broad sensory settings.
The highest-accuracy setting differs from the best-GCS setting, showing that recognition performance alone is insufficient for evaluating human-like scanpaths.
These findings motivate bias-aware evaluation practices for active vision and gaze-alignment benchmarks.

\section*{Acknowledgments}
We acknowledge the use of AI tools for language editing and draft refinement; all analyses, results, and claims were verified by the authors. This work was supported by the JST SPRING GX Program (Grant Number JPMJSP2108) and JSPS KAKENHI Grant Number JP25K24741. We thank Prof. Hirokazu Takahashi for insightful discussions and helpful comments that inspired this work.

\printbibliography

\end{document}